# Compact pneumatic clutch with integrated stiffness variation and position feedback

Yongkang Jiang[1,2,#], Junlin Ma[1,2,#], Diansheng Chen[1,2,*], and Jamie Paik[3,*]

*Abstract*— Stiffness variation and real-time position feedback are critical for any robotic system but most importantly for active and wearable devices to interact with the user and environment. Currently, for compact sizes, there is a lack of solutions bringing high-fidelity feedback and maintaining design and functional integrity. In this work, we propose a novel minimal clutch with integrated stiffness variation and real-time position feedback whose performance surpasses conventional jamming solutions. We introduce integrated design, modeling, and verification of the clutch in detail. Preliminary experimental results show the change in impedance force of the clutch is close to 24-fold at the maximum force density of 15.64 N/cm². We validated the clutch experimentally in (1) enhancing the bending stiffness of a soft actuator to increase a soft manipulator's gripping force by 73%; (2) enabling a soft cylindrical actuator to execute omnidirectional movement; (3) providing real-time position feedback for hand posture detection and impedance force for kinesthetic haptic feedback. This manuscript presents the functional components with a focus on the integrated design methodology, which will have an impact on the development of soft robots and wearable devices.

## I. INTRODUCTION

The ability to adjust structural stiffness is crucial for soft robots and wearable devices in circumstances where large exerting forces are required to manipulate objects or assist the wearers, or high mechanical stiffness is needed to block motions in a specific direction [1, 2]. In recent years, we have seen a number of solutions for modulating mechanical stiffness, including jamming mechanisms [3-7], phase change materials [8-13], and antagonistic structures [14, 15]. However, the range of stiffness change, miniaturization, and the sensing ability still challenge current variable stiffness mechanisms. Additionally, real-time position feedback has a critical role in posture estimation and precise control for soft robots and wearable devices. Research efforts have strived to obtain the sensing ability, such as conductive-fluid-based soft sensors [16, 17]. Nevertheless, in most of today's work, stiffness variation structures and position sensors are designed and even fabricated separately. Then, these individual parts are put together simply to form a multi-functional system, which leaves room for optimizing and streamlining the integration step [18]. Therefore, we need a new integrated design method to realize stiffness modulation and position feedback in one step for soft robots and wearable devices.

As core components of many robotic systems, clutches play a vital part in changing various motion types and allowing stiffness variation with the antagonistic arrangement method [19]. There has been wide use of conventional rigid clutches [20], but they are rather bulky for soft robots and wearable devices. The recently proposed soft textile clutches [21] and electroadhesive clutches [22] show great advantages due to their flexible bodies; however, most of the soft textile clutches utilize vacuum power for jamming to generate high impedance force [23]. Similar to the existing layer-jamming mechanisms [24, 25], the relatively low force density, high hysteresis, lack of sensing ability, and cumbersome bodies might be limitations for these jamming-based solutions. For electroadhesive clutches that utilize electroadhesion force to block a specific motion, the high driving voltage, possible electric breakdown, and remaining adhesion could be their shortcomings [26]. Furthermore, a significant force drop has been observed when the external pulling force exceeds the maximum blocking force in electroadhesive clutches [27, 28], which means they are more suited to changing motion type by blocking specific movements, rather than realizing continuous stiffness variation. Previously, we used soft silicone actuators to push together two high-friction layers and confirmed that friction-force-based clutches have merits in continual stiffness changing [29]. However, the low force density, as well as cumbersome bodies, restricted their further application. As a new type of soft actuator, air pouches made of several layers of flexible materials have shown considerable potential as they are ultralight, customizable, and easy to fabricate [30], making them a perfect alternative to the soft silicone actuators of our previous work. Nevertheless, most of the existing effort has focused on taking advantage of the in-plane contraction of air pouches, which hardly exceeds 36% [31]. In comparison, it has neglected its large deformation rate (x100) in the out-of-plane direction. Indeed, actual work usually adopted small deformation plate theory, ignoring the elongation in the middle plane to describe the shape function of the air pouches, which could induce a significant deviation between analytical and experimental results [32]. To better predict the relationship between the input air pressure and friction force of a possible air-pouch-based clutch, we need to analyze the large deformation of the air pouches more comprehensively.

Here, we propose a compact clutch with an air pouch that actively modulates the impedance force and integrating a sensor to provide real-time feedback. The presented clutch uses positive air pressure for a fast-response layer-jamming

The research is supported by the National Key R&D Program of China (2018YFB1304600), National Natural Science Foundation of China (51775012), Beijing Natural Science Foundation (L182012), and Beijing Municipal Science & Technology Project (Z191100004419006).
# The authors contribute equally to this work.
[1] Institute of Robotics, Beihang University, Beijing, 100191, China.
[2] Beijing Advanced Innovation Center for Biomedical Engineering, Beihang University, Beijing, China.
[3] National Research Center for Rehabilitation Technical Aids, Beijing, 100176, China.
* Corresponding authors: Diansheng Chen (chends@buaa.edu.cn), Jamie Paik (Jamie.paik@epfl.ch)

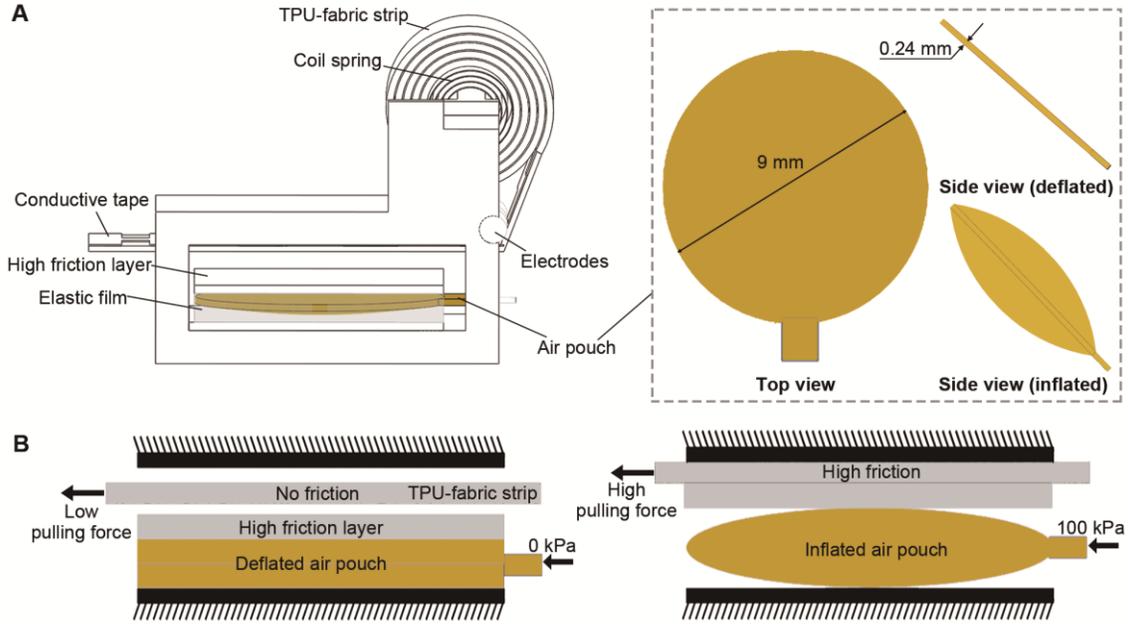

Figure 1. Schematic of the proposed clutch consisting of an air pouch for regulating the impedance force, a sensing layer for real-time position feedback, and a coil spring for recovery of the TPU-fabric strip. (A) System overview. (B) Working principle of the air pouch part.

for high impedance force enhancement, increasing the overall jamming force compared to the state of the art, which can only apply that maximum of negative vacuum pressure (85-90% vacuum). With this novel unified design, the above shortcomings of the conventional layer-jamming mechanisms such as high hysteresis can be avoided, and at the same time, a wide range of impedance force adjustment and sensing ability are achieved. We first describe the structure and working principle of the planned clutch to feature the integrated design method in this work. As a central part of the clutch, we provide a mathematical model of the air pouch in two layers based on a large deformation plate theory that explores the deformation-pressure response. Additionally, we present the relationship between the normal pushing force of the air pouch and the inside air pressure in detail to predict the friction force. We then verify the theoretical models experimentally. Other tests analyzed influence factors on the impedance force and sensing ability of the clutch. Finally, we evaluated the performance of the proposed clutch with three functional prototypes to highlight its potential applications in soft robots and wearable devices, including in (1) enhancing the grasping force of a soft gripper by modulating its bending stiffness; (2) enabling omnidirectional movement of a soft cylindrical actuator by selectively adjusting input air pressure inside the clutches; (3) providing real-time position feedback for hand posture detection and impedance force for possible kinesthetic haptic feedback in virtual reality (VR) applications.

The main contributions of this work are:

- Integrated design of a novel compact clutch with modulable stiffness and real-time position feedback;
- Modeling, validation, and characterization of the original high force clutch design;
- Performance evaluation of the clutch in three potential applications, including a soft gripper, a multi-DoF actuator, and a hand-motion tracker.

## II. A NOVEL COMPACT CLUTCH WITH MODULABLE STIFFNESS AND POSITION FEEDBACK

Stiffness variation and real-time position feedback are pivotal in the design of a clutch. The main difficulty in enabling stiffness variation based on the antagonistic arrangement method is changing their impedance force over a broad-enough range. Furthermore, miniaturization of the clutch to make it suitable for soft robots and wearable devices is another obstacle. Here we present the system overview and working principle of a new clutch using an integrated design method to overcome these problems. A comprehensive model of the large deformation of the air pouch and the force-pressure response of the clutch can be found in the Appendix.

We illustrate the clutch, made up of three main parts, in Fig. 1A. Parts include an air pouch modulating the friction force on a TPU-fabric strip, a sensing layer detecting displacement and movement direction of the TPU-fabric strip, and a coil spring for recovering the TPU-fabric strip. Unlike other systems where stiffness variation and sensing parts have no common elements and work independently, in this design the TPU-fabric strip with a conductive tape can produce a high friction force when the air pouch pressure increases, and provide position feedback when the strip comes out of the clutch. This combined system not only reduces the size of the whole structure but also the mechanical interference between the layers.

The air pouch, also shown enlarged Fig. 1A, provides a normal pushing force generating large changes in impedance force. A high friction layer is utilized to avoid direct contact of the air pouch with the TPU-fabric strip to enhance durability. Likewise, elastic films promote fast recovery when the air pouch is deflated. The basic working principle of the clutch for regulating the impedance force on the TPU-fabric strip is illustrated in Fig. 1B: when the air pouch is deflated, the TPU-fabric strip moves freely with low impedance force; Once the air pouch is inflated, the high friction layer and the frame of the clutch block the strip, significantly increasing the friction force. Then, the impedance force can be continuously regulated by changing the air pressure inside the air pouch. Of note, the ground (black hatched lines on top and bottom) in Fig. 1B refers to the 3D printed frame of the clutch. Besides, the elastic film is not shown in Fig. 1B for clarity.

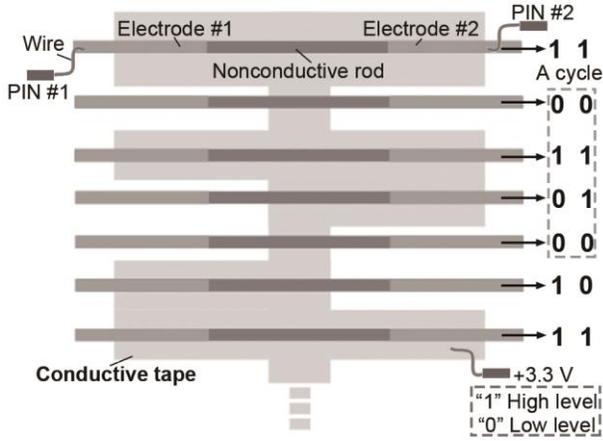

Figure 2. Basic concept of the sensing part that communicates the displacement and moving direction of the TPU-fabric strip.

The main components and the general sensing principles are given in Fig. 2. A piece of super-thin (0.06 mm) and flexible conductive tape is designed of specific patterns and fabricated using a 2D manufacturing method [31] [33, 34]. The conductive tape is then attached to the top surface of the TPU-fabric strip so that it could follow the movement of the strip and wrap around the coil spring. At the back of the printed frame as in Fig. 1A, we utilized a cylindrical non-conductive rod to guide the movement of strip. Notably, there are two pieces of copper foil wrapped around the non-conductive rod at both ends that act as electrodes. As shown in Fig. 2, we permanently connected the conductive tape to +3.3V on an Arduino board, and linked the two electrodes on the rod to different interface ports, respectively. When the electrode is not in contact with the conductive tape, a low-level output can always be detected at the related port due to the pull-down resistance (not shown in the figure for simplification). Similarly, we can obtain a high-level output when the electrode contacts the conductive tape. As the TPU-fabric strip moves under an external pulling force, the electrodes on the non-conductive rod will contact or lose contact with the conductive tape in turns, and we give specific sequences of status in Fig. 2. When the sequence changes once, it means that the strip moves forwards or backwards for a prescribed distance (1mm in this work, which is also the designed resolution of the sensing part). Similar to the working principle of traditional incremental encoders, we can obtain the displacement, as well as the moving direction of the strip, by counting the number of times that the status of sequences changed and analyzing the order between the current status and former one. Especially, we designed a specific pattern at the beginning of the conductive tape for calibrating and indicating the initial position at which the displacement of the strip is zero.

Here, we used the elastic film made of a 0.06 mm adhesive TPU, and a TPU-fabric strip laser-cut from 0.2 mm fabric-coated with TPU films on both sides. Also, we bonded two layers of 0.12 mm fabric coated with TPU films on one side together to form the air pouch. Of note, we adopted circular air pouches in this work for simplification, while air pouches with other shapes could also work, which we will investigate in the future.

III. RESULTS

First, we compared analytical results with simulational and experimental ones to verify the theoretical models as given in the appendix. Additionally, we evaluated the performance of the clutch experimentally, including how the impedance force changes as the displacement of the TPU-fabric strip increases for different input air pressures, and the influence factors on the sensing ability of the clutch. Furthermore, we also conducted several groups of fatigue life tests on the clutches to study the endurance after a certain number of working cycles. Finally, we utilized the proposed clutch in three potential applications to evaluate its effectiveness.

A. *Verification of the air pouch models*

We conducted standard uniaxial extension tests of the utilized flexible materials, including TPU-fabric material for the air pouches and adhesive TPU material for the elastic films, to obtain essential material coefficients before simulation and experimental verification of the theoretical models. For the TPU-fabric material, we found that the nominal stress-strain curve was almost linear, and we obtained the corresponding elastic module as $E = 142.2\ MPa$. For the adhesive TPU material, the coefficients for the third-order Ogden material model were given as $u_1 = -16.47; \alpha_1 = 1.61; u_2 = 6.47; \alpha_2 = 2.30; u_3 = 13.22; \alpha_3 = 0.71$ with MPa units.

First, we verified the theoretical model concerning the deformation-pressure relationship of the air pouch with both FEA simulation and experiments. Due to the symmetry, we analyzed only half of the air pouch noted as the thin plate in the aforementioned model with ABAQUS software to save computation costs. The boundary conditions we used in the simulation were identical to reality. The simulation results showed that the thin plate would deform into a dome-like shape when inflated, and the maximum deflection happened at the center of the plate as the theoretical model predicted. Then, we extracted the maximum deflections under different air pressures for further comparison between analytical and experimental results.

In the experiments, we tested three air pouches with identical geometry to obtain the maximum deflections as the air pressure increased. Experimental, analytical, and simulational results are shown in Fig. 3. As depicted, the deviation among the results was acceptable within the error margin. Here, the curves tended to diverge under relatively high air pressure, where a large deformation occurred. According to reference [35], the equation describing the radial strain of the plate has to be modified in cases of very large deflections, which might be one of the causes inducing the deviation.

As mentioned above, the inflated air pouch will push up the FR4 layer to increase the normal pushing force acting on the TPU-fabric strip. To verify the theoretical model, we tested three air pouches to obtain the normal pushing force with an increase in the pressure. In this test platform, we mounted a force gauge above the air pouch with a gap identical to the real distance between the top face of the air pouch part and the TPU-fabric strip in the clutch. During the experiments, we inflated the air pouches to contact the force gauge, and the normal pushing forces as the air pressure increases were recorded. As illustrated in Fig. 4, the experimental and analytical results matched well considering fabrication and measurement errors.

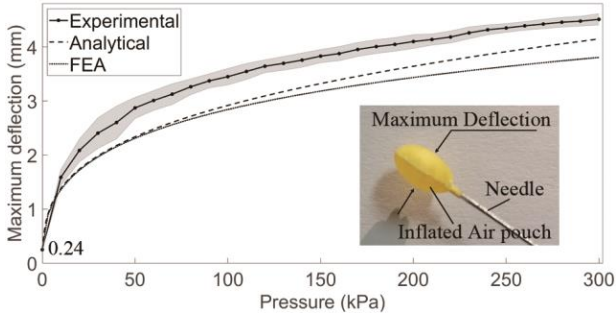

Figure 3. Experimental, analytical, and FEA results of the maximum deflection at the center of the air pouch when the input air pressure varies.

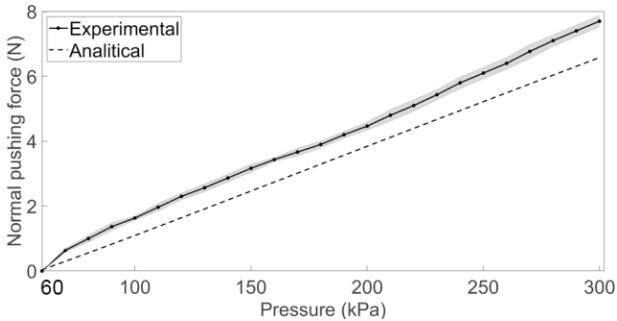

Figure 4. Experimental and analytical results of the normal pushing force that the air pouch module exerts on the TPU-fabric strip.

### B. Performance evaluation of the clutch

To describe completely the performance of the proposed compact clutch, we conducted several groups of experiments to evaluate its force-pressure response and sensing ability.

When the pulling force acting on the TPU-fabric strip exceeds the impedance friction, the strip will be pulled out of the clutch gradually, which means the displacement of the strip increases. We conducted tests on a clutch for three times when the air pressure inside the air pouch was set at 0 kPa, 100 kPa, 200 kPa, and 300 kPa, respectively. As depicted in Fig. 5, for higher input air pressure, a larger pulling force was needed for the strip to move. At the initial position, where the displacement of the strip is zero, the pulling force was 0.53 N at 0 kPa pressure and 12.67 N when the pressure was 300 kPa, which means it achieved about a 24-times pulling-force enhancement. Furthermore, the force density of the clutch at 300 kPa input air pressure was 15.64 N/cm$^2$, which is of the same magnitude as the recently reported high force density electrostatic clutches as in [27], and much higher than other jamming-based solutions [21] [23-25]. To note, all the curves as shown in Fig. 5 have an upward tendency because the recovery force of the customized coil spring of Fig. 1A gets larger as the displacement of the strip increases. In addition, a possible cause for the fluctuation on the curves may be the elasticity of the TPU-fabric strip and the difference between the maximum static friction and dynamic friction forces.

Then, we characterized the sensing ability of the proposed clutch. As mentioned before, the two electrodes on both ends of the non-conductive rod will contact the conductive tape on the top of the TPU-fabric strip alternately when the strip is pulled out of the clutch, and hence we can obtain four sequences of status by specific turns to calculate the displacement of the strip and detect of its moving direction. Notably, when we pulled out the strip, we define it as the "forward" moving direction throughout this paper. Otherwise, it is defined as "backward". Experiments were then conducted to study the influence of the moving speed of the strip and air pressure inside the air pouch on the sensing ability of the proposed clutch. As illustrated in Fig. 6, the experimental results matched well with the corresponding reference curves when we pulled out the strip of the clutch at a speed of 2 mm/s, 1 mm/s, and 0.2 mm/s, respectively. The maximum deviation between the experimental and theoretical results on the displacement of the strip did not exceed 1 mm, which corresponded to the designed resolution of the sensing part: this confirmed that the sensing part of the clutch worked well within the particular moving speed range.

Figure 7 demonstrates further the effectiveness of the sensing layer when the air pressure in the air pouch varies. As can be seen, the experimental results matched well with the reference curve, especially for relatively low air pressure. When the air pressure increased, the deviation between the experimental and reference results tended to ascend. A possible reason is that the shape of the conductive tape attached to the top of the elastic TPU-fabric strip could be pulled and changed for higher input air pressure where a larger pulling force was needed for the strip to move. However, the clutch was always utilized as a blocking mechanism where the strip hardly moved when the air pouch was inflated at 300 kPa. Hence, the "harmful" situation for the sensing part as depicted in Fig. 7D rarely happens in real applications.

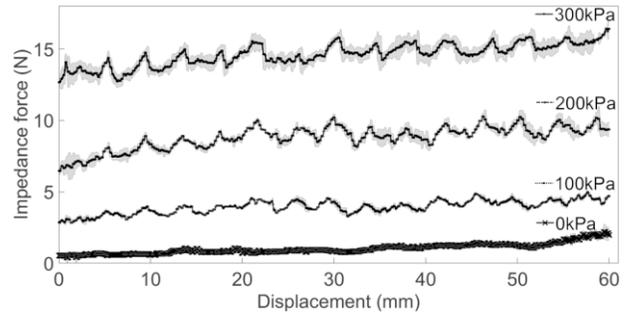

Figure 5. Impedance force of the TPU-fabric strip as the displacement of the strip increases while the air pressure in the air pouch is set at 0 kPa, 100 kPa, 200 kPa and 300 kPa, respectively.

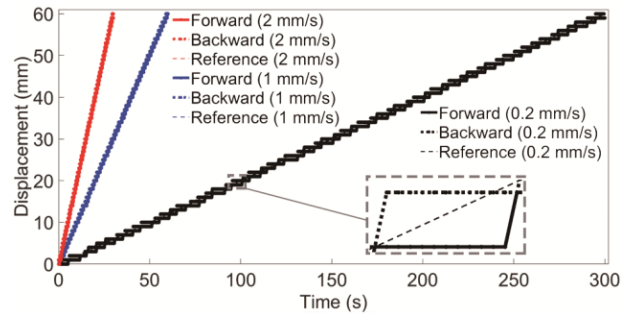

Figure 6. Evaluation of the sensing ability of the minimal clutch when the TPU-fabric strip is pulled out at a speed of 2 mm/s, 1 mm/s and 0.2 mm/s, respectively.

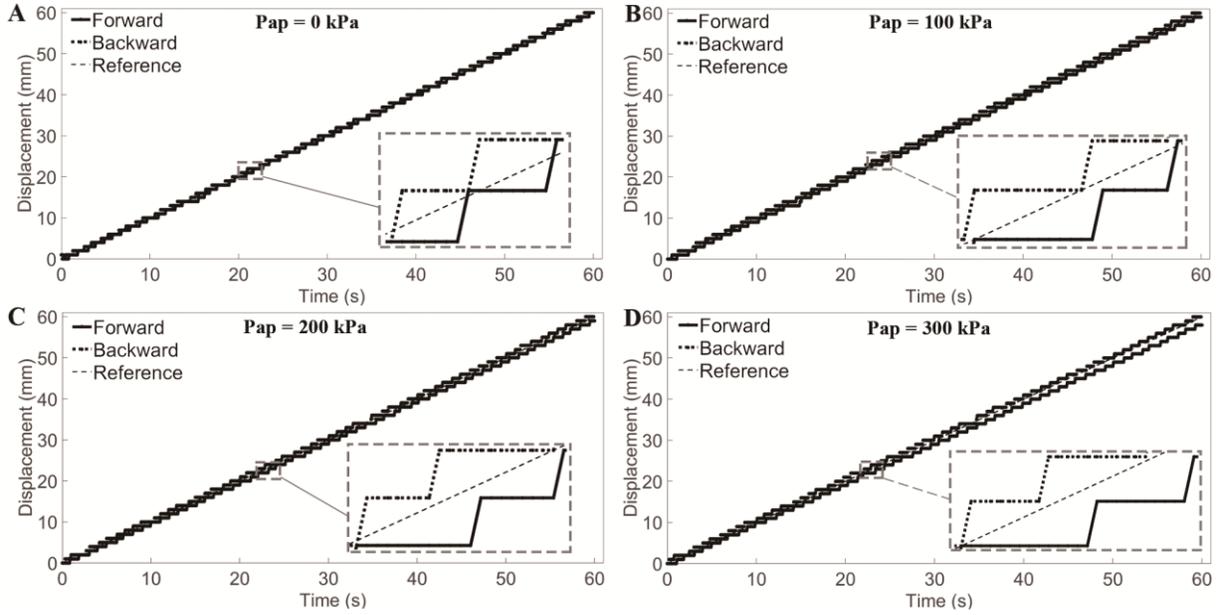

Figure 7. Evaluation of the sensing ability of the minimal clutch when the air pressure inside the air pouch was set at 0 kPa, 100 kPa, 200 kPa and 300 kPa respectively, and the TPU-fabric stirp was pulled out at a speed of 1 mm/s.

Furthermore, we did fatigue life tests to evaluate the endurance of the clutch when the air pouch was inflated with different input air pressures. Details are given in the appendix. Preliminary results showed that after 100 working cycles (determined experimentally), the prototyped clutch still performed well under relatively low air pressure. However, a gradual weakening trend was observed when the air pressure increased.

These results show that the proposed clutch not only regulated the impedance force in a wide range (~x24) and provided precise position feedback within 1 mm resolution, but also remained functional after repeated working cycles.

*C. Potential applications*

To further demonstrate the effectiveness of the new clutch, we also investigated its performance in three potential applications, including (1) increasing the bending stiffness of a soft actuator, and gripping force of a soft gripper; (2) realizing omnidirectional movement of a soft cylindrical actuator by selectively modulating input air pressure of the clutches; (3) providing real-time position feedback for finger posture detection and impedance force for possible kinesthetic haptic feedback in VR applications.

As illustrated in Fig. 8(A-1), we mounted the pneumatic clutch on the proximal end of a bending soft actuator and attached the tip of the TPU-fabric strip of the clutch to the distal end. A forcemeter installed on a translational platform was connected to the distal end of the actuator via an inextensible cable. When the forcemeter was driven by the platform to move away from the soft actuator, we recorded the horizontal displacement of the distal end of the soft actuator and the corresponding pulling force $F_{Ta}$. Then, we could easily obtain the bending stiffness of the soft actuator when it was in the straight state. In the experiments, we pulled the distal end of the soft actuator to have a 10 mm displacement, which we chose empirically when the air pressure inside the soft actuator was set at 0 kPa. The bending stiffness of the soft actuator in this situation was $5.5 \times 10^{-3}$ N/mm. Then, the clutch was inflated with 300 kPa air pressure. Experimental results showed that the soft actuator could maintain a straight state until the air pressure inside it exceeded 200 kPa. To obtained the maximum bending stiffness of the soft actuator in the straight state, we set the air pressure of the clutch and the soft actuator at 300 kPa and 200 kPa, respectively. Then, the distal end of the actuator was also pulled for 10 mm. With a similar calculation method, the maximum bending stiffness was $3.8 \times 10^{-2}$ N/mm, which is about 7 times the stiffness when the clutch and soft actuator were inactive.

To better show the significance of this stiffness enhancement, we proposed an experimental platform as depicted in Fig. 8(A-2) to test the gripping force of a soft gripper made of two identical soft actuators as shown in Fig. 8(A-1). First, we removed the clutches mounted on the soft actuators and inflated the actuators to completely wrap the 3D-printed cylindrical object. These experimental results showed that too large an air pressure would induce excessive deformation of the soft actuators [37, 37] to lose contact with the object, and lead to grasping failure. Hence, the air pressure inside the soft actuators was set at 100 kPa accordingly to avoid excessive bending and provide the maximum gripping force for the soft gripper without the clutches. In the experiment, the cylindrical object was pulled up slowly by a forcemeter. The results showed that the maximum pulling force $F_{Tg}$ throughout the whole test was 3.97 N (the weight of the object has been compensated in these experiments). Then, we re-mounted the clutches back onto the soft gripper and tested the maximum gripping force. When the clutches were inflated at 300 kPa, we found that the maximum air pressure before the undesired deformation happened was 200 kPa. Similarly, we obtained the maximum gripping force of the soft gripper with the proposed clutches of 6.88 N, which is 73.3% higher than the gripper without clutches.

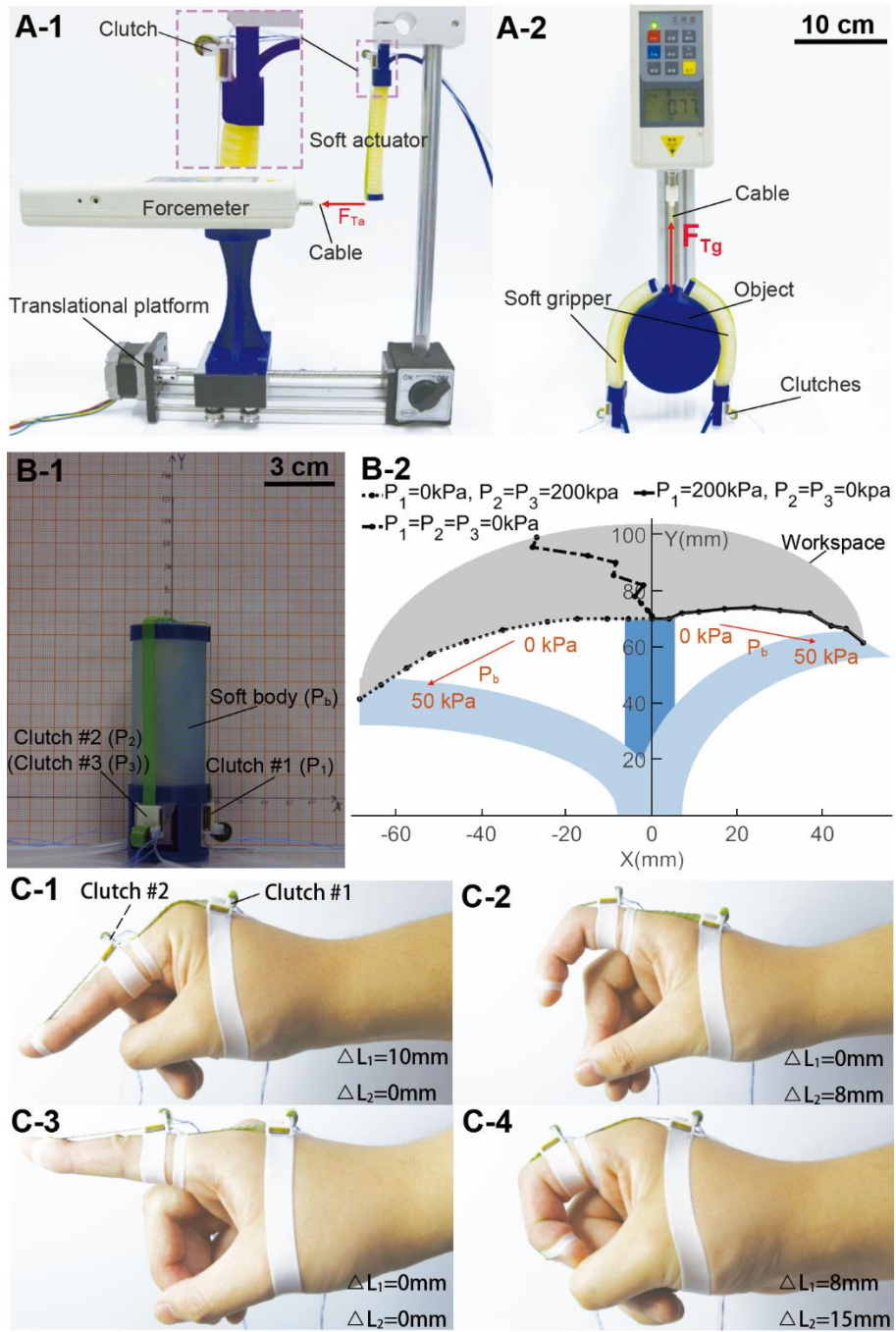

Figure 8. Performance evaluation of the clutch in three potential applications. (A) Enlarging the gripping force of a soft gripper by enhancing the bending stiffness; (B) Enabling omnidirectional motion of a soft cylindrical actuator by selectively adjusting the air pressure inside the clutches; (C) Providing real-time position feedback for hand posture detection.

Another potential application for the proposed minimal clutch is to enable multi-degrees-of-freedom (DoF) motion of soft actuators by selectively restricting deformation in specific directions. As illustrated in Fig. 8(B-1), we evenly distributed three identical clutches on the proximal end of a soft cylindrical actuator, while we mounted the tip of their strips on the distal end. Experimental results showed that the soft actuator was able to achieve omnidirectional motion by selectively adjusting the air pressure inside of the three clutches when the soft actuator was inflated with different pressures. For example, the actuator just elongated upwards when no clutch was inflated. However, the actuator bent towards clutch #1 when the air pressure inside clutch #1 was larger than that of clutch #2 and #3, and vice versa. For simplification, we only recorded only the motion in the XY plane. As can be seen in Fig. 8(B-2), each mark refers to a position of the distal end of the soft actuator under different circumstances. All the marks made up the workspace of the soft actuator when the air pressure inside the actuator varied from 0 kPa to 50 kPa, and the pressure inside the three clutches changed within 0-200 kPa.

Finally, we evaluated the proposed clutch on a human hand to show its advantages for miniaturization, in real-time position sensing, and possible kinesthetic haptic feedback for VR applications. As depicted in Fig. 8C, we put two prototyped clutches on an index finger. Clutch #1 was bonded onto the metacarpal bone and the tip of its TPU-fabric strip to the proximal phalanx to measure the motion of the metacarpophalangeal (MCP) joint. Similarly, clutch #2 was bonded on the proximal phalanx and the tip of its TPU-fabric strip to the distal phalanx for sensing of the proximal interphalangeal (PIP) and distal interphalangeal (DIP) joints. When the joints of the index finger moved, the relative displacement of the strips $\Delta L_1, \Delta L_2$ in clutch #1, and #2, respectively changed as can be seen in Fig. 8 (C-1) to (C-4). Then, by calibration and simple calculation, we could easily obtain the bending angles of each joint. Hence, the posture of the index finger could be further detected, and the method can be easily expanded to the whole human hand or even other wearable devices.

The demonstrations introduced above are just some typical applications of the projected minimal clutch to show its effectiveness. With the above merits such as fast response, high force density, miniaturized body, and real-time displacement feedback, the clutch could contribute to various working scenarios. For instance, the minimal clutch showed its advantages in adjusting the bending stiffness of continuum soft actuators of diverse shapes in specific directions to enhance the gripping force or carry out multi-DoF motions as illustrated in Fig. 8(A-B). In addition, for articulated rigid robots with multiple joints, the clutch could also help in increasing the structural stiffness of the joints to sustain higher external loads or execute various motions.

Furthermore, the new clutch could also provide customized impedance force at each joint of the human hand to imitate kinesthetic haptic feedback when the user contacts objects of different hardness or grasps objects with varied weight in VR applications as described in [27].

IV. CONCLUSION AND FUTURE WORK

In this paper, we present a novel design of a compact pneumatic clutch for soft robots and wearable devices accomplishing integrated stiffness variation and position feedback. First, we introduced the integrated design of the minimal clutch utilizing an air pouch to adjust continuously the impedance force and an ultra-thin patterned conductive trace to provide real-time position feedback. We also presented theoretical models to study in detail the large deformation of the air pouch and force-pressure response of the clutch. The experimental, simulational, and analytical results matched well within a reasonable error margin. We then tested the performance of the proposed clutch under different conditions. Preliminary results showed that the force density of the clutch at 300 kPa input air pressure was 15.64 N/cm2, much higher than other jamming-based clutches.

Furthermore, the impedance force on the TPU-fabric strip of the clutch could be enhanced about 24 folds when the air pressure inside the air pouch increased from 0 kPa to 300 kPa. For sensing, results showed that the maximum deviation between the experimental and theoretical results for the displacement of the strip did not exceed 1 mm when the strip was pulled out of the clutch within a moving speed range of 0.2 mm/s - 2 mm/s, and the air pressure inside the air pouch varied from 0 kPa to 200 kPa.

Finally, we also investigated the proposed clutch in three potential applications to further demonstrate its effectiveness. Experimental results show that (1) the clutch could enhance the bending stiffness of a soft actuator up to 7 times, and enlarge the gripping force of a soft gripper up to 73.3%; (2) a soft cylindrical actuator could carry out omnidirectional movement by selectively adjusting the air pressure inside the actuator and three clutches; (3) the clutches placed on human fingers could provide real-time position feedback for posture detection and impedance force for possible kinesthetic haptic feedback in VR applications.

In the future, we will optimize the design of this new clutch, including testing air pouches of varying geometry to further improve the force density and utilizing new materials for the conductive tape to increase the fatigue life. Also, we will improve the flexibility of the overall clutch for use in wearable applications.